\documentclass[runningheads]{llncs}
\usepackage[T1]{fontenc}
\usepackage{graphicx}
\usepackage{hyperref}
\usepackage{multirow, tabularx, booktabs}
\usepackage{amsmath, amssymb}

\usepackage{appendix}
\begin{document}
\title{Selective Time Series Forecasting via Metalearning
}
\titlerunning{Selective Time Series Forecasting via Metalearning}
%

\author{Ricardo~Inácio\inst{1}\orcidID{0009-0008-6435-5245}, Vitor~Cerqueira\inst{2}\orcidID{0000-0002-9694-8423}, Marília~Barandas\inst{3}\orcidID{0000-0002-9445-4809} \and Carlos~Soares\inst{1,3}\orcidID{0000-0003-4549-8917}}

\authorrunning{R. Inácio et al.}

\institute{Faculdade de Engenharia da Universidade do Porto, Porto, Portugal \\
\email{\{rcinacio,csoares\}@fe.up.pt} \and
University of Coimbra, Portugal\\
\email{vitorc@dei.uc.pt} \and
Fraunhofer Portugal AICOS, Portugal\\
\email{marilia.barandas@aicos.fraunhofer.pt}
}

\maketitle      

\begin{abstract}

Deep learning methods have achieved state-of-the-art in time series forecasting, yet their accuracy varies considerably across samples, as some instances remain inherently difficult to predict. Reject option mechanisms, which allow models to abstain from high-risk predictions, are well established in classification and regression but underexplored in forecasting. Existing abstention strategies typically rely on proxies, such as the width of the prediction interval or learned confidence scores derived from forecasts. However, these approaches are inherently tied to the training domain, limiting their ability to generalize.
We propose a selective forecasting framework that addresses this limitation by modeling the empirical percentile of forecasting errors, that is, a scale-invariant statistic, based on structural characteristics extracted from recent lags via metalearning. By decoupling the rejection decision from the forecast itself and grounding it in domain-agnostic features, the framework enables effective abstention transfer across heterogeneous time series. Experiments in both in-domain and transfer learning settings show that rejecting samples predicted as challenging consistently improves forecasting accuracy across coverage levels.

\end{abstract}

\setcounter{footnote}{0} 
\section{Introduction}
\label{sec:intro}

\begin{figure}[t]
    \centering
    \includegraphics[width=1\textwidth]{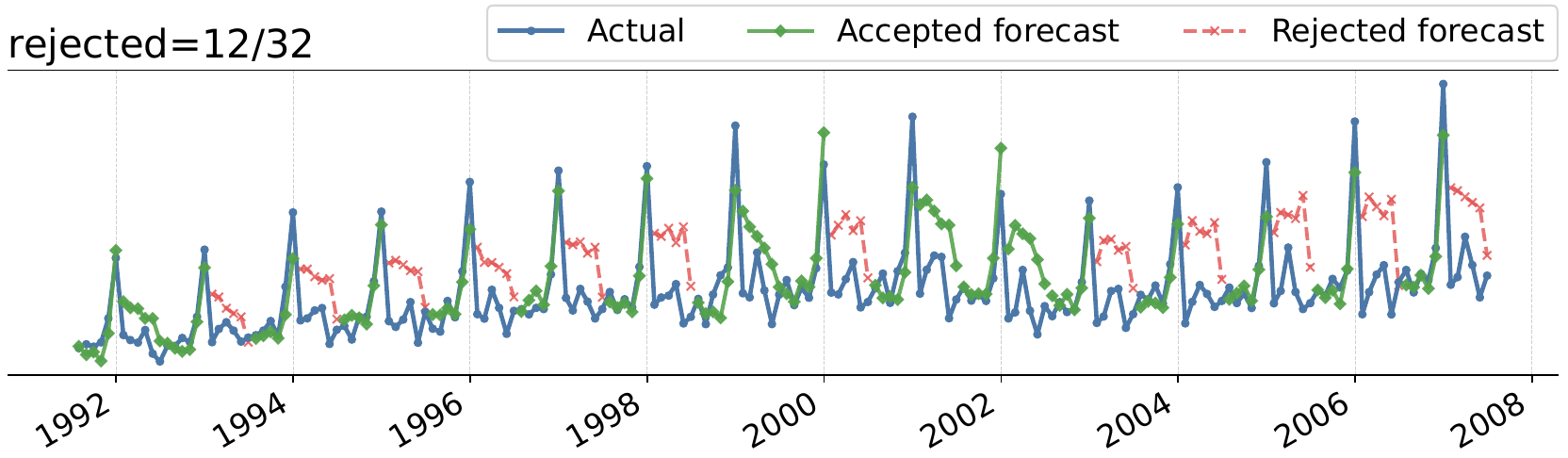}
    \caption{Example application of the proposed method for selective forecasting. A metamodel scores each forecast origin, accepting predictions expected to be reliable (green) and rejecting those likely to incur large errors (red). Rejected forecasts occur during the most challenging periods, when predictions deviate substantially from observations.}
    \label{fig:cumulative_forecast}
\end{figure}

Deep learning methods have become the dominant approach in time series forecasting, achieving state-of-the-art results across diverse domains~\cite{aksu2024gifteval}. Yet their accuracy is not uniform: certain samples remain inherently difficult to predict, and errors can vary substantially across a time series~\cite{aiolfi2006persistence}. This motivates reject-option mechanisms, which allow models to abstain when risk is high. Despite the non-uniform nature of forecasting errors, such mechanisms remain underexplored in time series compared with classification~\cite{chow1970} and regression~\cite{shah2022selective}. Effectively identifying hard-to-predict periods in advance would allow practitioners to selectively avoid unreliable predictions.

Existing approaches to abstention in forecasting typically rely on proxy heuristics derived from forecasts, such as prediction interval width~\cite{szabadvary2025classification}, nonconformity scores~\cite{vovk2005algorithmic,stankeviciute2021conformal}, or learned confidence estimates~\cite{brusokas2025time}. However, these signals are inherently tied to the trained model and its domain: prediction intervals reflect model-specific uncertainty calibration, and confidence scores are learned on the same data distribution used for forecasting. As a result, such rejection mechanisms do not generalize well when applied to new domains or time series with different characteristics.

To address this limitation, we propose a selective forecasting framework based on metalearning. The key idea is to model the empirical percentile of forecasting errors, that is, a scale-invariant statistic, based on structural features extracted from recent lags. Because these features summarize local patterns (e.g., trend, seasonality, lumpiness) rather than forecast outputs, and because percentile ranks are normalized across series, the rejection mechanism is decoupled from both the forecasting model and the data scale. This design enables the metamodel to transfer effectively across heterogeneous domains.

We train on a source domain by computing error scores and assigning them empirical percentile ranks, which serve as regression meta-targets. Time series features extracted from recent lags~\cite{barandas2020tsfel,cerqueira2024vest} serve as inputs. At inference, the metamodel outputs a continuous score for each forecast origin, rejecting those exceeding a predefined threshold $q$ (Figure~\ref{fig:cumulative_forecast}). 


We evaluate the framework in both in-domain and transfer learning settings, showing consistent forecasting accuracy gains across coverage levels. In summary, our contributions are: i) a metalearning approach for large forecasting errors estimation using structural properties of recent lags and scale-invariant targets; and ii) empirical evidence that the approach transfers effectively across domains, outperforming forecast-derived heuristics. Code and data are available online.~\footnote{\href{https://github.com/ricardoinaciopt/selective_forecasting_metalearning}{https://github.com/ricardoinaciopt/selective\_forecasting\_metalearning}}
\section{Related Work}

\subsection{Time Series Forecasting}
\label{sec:forecast}

Let $\mathcal{Y} = \{ Y^{i} \}_{i=1}^n$ denote a collection of $n$ time series, where each \( Y^{i} = \{ y^{i}_j \}_{j=1}^{n_i} \) is a sequence of observations indexed by time $j$, and $n_i$ is the length of the $i$-th series. Global forecasting models train a single model on $\mathcal{Y}$ exploiting shared structure across series, often improving accuracy and data efficiency relative to local (i.e., per-series) models~\cite{januschowski2020criteria,godahewa2021ensembles}. Using a context window of length $p$, each series is turned into input-output pairs via time-delay embedding~\cite{bontempi2012machine}. At each time step $t$, the input is the window of lags $X^{i}_t \in \mathbb{R}^p$ and the output are the subsequent $h$ observations (i.e, the horizon) $Y^{i}_{t+1:t+h} \in \mathbb{R}^h$, following: $X^{i}_t = \{ y^{i}_{t-p+1}, \dots, y^{i}_t \}, \qquad  Y^{i}_{t+1:t+h} = \{ y^{i}_{t+1}, \dots, y^{i}_{t+h} \}$.

Any time step $t$ at which a forecast is made is denoted as forecast origin~\cite{hewamalage2023forecast}, where the next $h$ points are predicted from their previous $p$ lags. Accordingly, training amounts to fitting a global function $f_\theta: \mathbb{R}^p \rightarrow \mathbb{R}^h$ so that $f_\theta(X^{i}_t) = \hat{Y}^{i}_{t+1:t+h}$ approximates the true future values across all series and time steps.

Deep learning offers a wide variety of such approaches to time series forecasting, including recurrent, convolutional, attention-based, and feed-forward designs, such as \texttt{NHITS}~\cite{challu2023nhits} or \texttt{KAN}~\cite{liu2025kan}, which have been able to attain high performance and efficiency, leading to favorable results in forecasting benchmarks and competitions, for instance, GIFT-Eval~\cite{aksu2024gifteval}.

\subsection{Reject Option and Selective Predictions}
\label{sec:reject_option}

\subsubsection{Traditional Reject Option.}

Reject-option introduces an abstention mechanism into predictive systems, allowing the model to withhold predictions when the expected cost of making a decision is high. Chow's rule formalizes this idea as an optimal decision strategy under a cost-sensitive loss: given posterior class probabilities, the classifier predicts the most probable class only when its confidence is sufficiently high, rejecting it otherwise~\cite{chow1970}. This induces an explicit trade-off between coverage (i.e., the fraction of accepted predictions) and risk (i.e., the error on the accepted subset).

This principle is commonly generalized under the selective prediction framework. 
Given a predictor $f$ and a selection function $s$ that outputs $1$ (accept) or $0$ (reject), the selective predictor is defined as:
$$
(f,s)(x) =
\begin{cases}
f(x), & \text{if } s(x)=1,\\
\bot, & \text{if } s(x)=0,
\end{cases}
$$
where $\bot$ denotes abstention. Following this, coverage is the fraction of instances that are accepted, $\phi(s)=\mathbb{P}[s(X)=1]$, and risk is the average loss over the accepted instances, $R(f,s)=\mathbb{E}[\ell(f(X), Y)\mid s(X)=1]$. The goal is to minimize risk while maintaining sufficient coverage, or equivalently, to maximize coverage subject to a target risk~\cite{el2010foundations,geifman2017selective}.

In the forecasting setting of Section~\ref{sec:forecast}, the predictor is the global model $f_\theta$, each instance corresponds to a forecast origin $t$ with input $X^{i}_t$ and target horizon $Y^{i}_{t+1:t+h}$, where the selection function $s(X^{i}_t)$ decides whether to issue the forecast $\hat{Y}^{i}_{t+1:t+h}$. The selective forecaster thus abstains at origins where the predicted error percentile is high, reducing average error at the cost of issuing fewer forecasts.

Performance is typically evaluated using risk-coverage curves, which report the mean error on the accepted instances at each coverage level, as progressively more instances are rejected (i.e., with less coverage). Moreover, AUCO (Area Under the Confidence-Oracle Error), defined as the distance between a risk-coverage curve induced by a given model and the oracle curve obtained by ranking instances by their realized errors, and ErrDrop, defined as the ratio of the no-rejection risk to the risk at the most selective coverage level, are other usual metrics that succinctly describe the error reduction achieved~\cite{scalia2020evaluating}.

\subsubsection{Selective Classification and Regression.}

In classification settings, rejection is often driven by confidence scores, calibrated probabilities, ensemble disagreement, or learned selection functions~\cite{geifman2017selective}. In regression tasks, there is no discrete class decision, so the relevant quantity is the expected error. The loss is therefore continuous, and rejection corresponds to withholding predictions whose estimated conditional error, variance, or uncertainty is high~\cite{wiener2012pointwise,shah2022selective}. 

A selective regression model can therefore be understood as a predictor equipped with a real-valued rejection score $s(x)$, where larger values indicate higher expected error. For a threshold $q$, predictions are rejected when $s(x)\geq q$. Stricter $q$ thresholds reduce the number of issued predictions, but should also reduce the average error among the accepted predictions~\cite{geifman2017selective,wiener2012pointwise}.

\subsubsection{Selective Forecasting.}

Most forecasting methods issue predictions continuously, without mechanisms to abstain when confidence is low~\cite{feng2025towards}. Some approaches provide uncertainty estimates, prediction intervals~\cite{szabadvary2025classification} or nonconformity scores~\cite{vovk2005algorithmic,stankeviciute2021conformal}, but these are designed for uncertainty quantification, not rejection. While interval width or nonconformity scores can serve as post-hoc rejection proxy heuristics, uncertainty is not equivalent to forecasting risk: wide intervals do not necessarily imply large errors, and narrow intervals may still occur in systematically missed regimes.

Recent work has introduced explicit selective forecasting mechanisms. Feng et al.~\cite{feng2025towards} propose rejection by ambiguity (based on error variance across training samples) and rejection by novelty (using VAEs to detect distribution shift). The Time-Energy Model~\cite{brusokas2025time} learns a confidence score jointly with the predictor to enable rejection under a risk-coverage trade-off. However, these methods rely on learned confidence or energy scores as proxies for risk rather than directly modeling the factors that cause large forecasting errors. Moreover, they remain tied to specific model architectures and training domains, limiting their applicability in transfer settings where rejections must generalize beyond the source data.

\section{Methodology}

\begin{figure}[h]
    \centering
    \includegraphics[width=\linewidth]{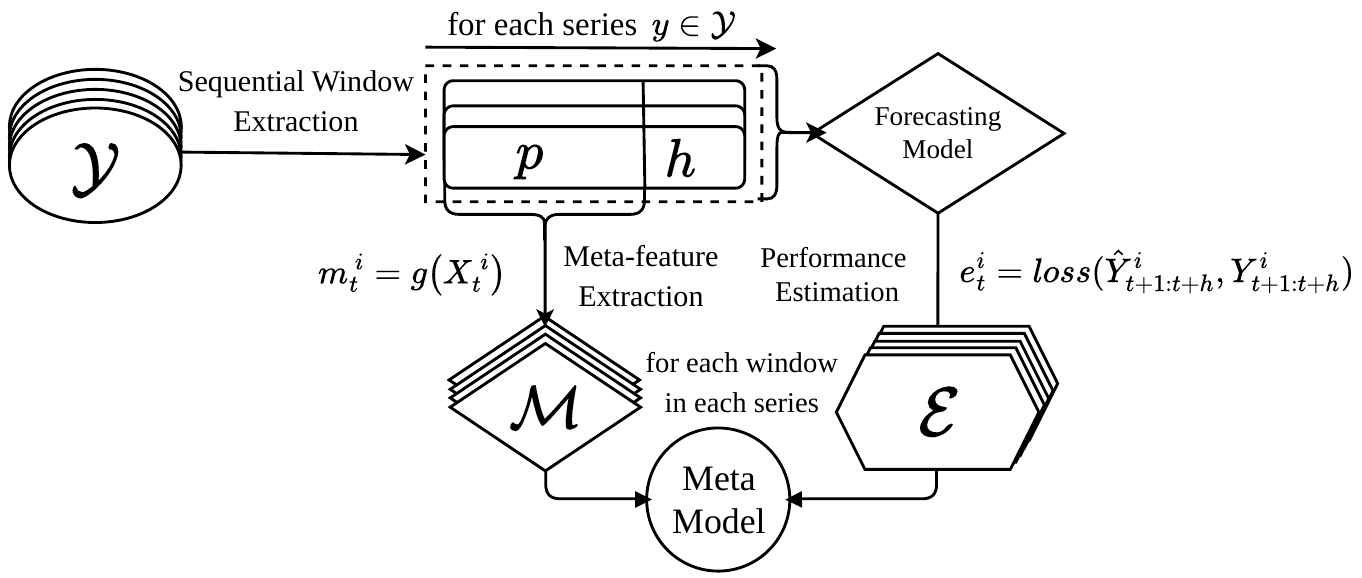}
    \caption{Overview of the proposed methodology. From each time series, a sequence of time-ordered windows is extracted, each composed of $p$ lags, the forecast origin, and the horizon $h$. Features are extracted from the lags, and performance estimates are obtained by measuring forecasting errors across each horizon. The metamodel is trained on the resulting predictors ($\mathcal{M}$) and targets ($\mathcal{E}$).}
    \label{fig:methodology}
\end{figure}

Our goal is to predict, before issuing a forecast, whether a global forecasting model is likely to incur a large error at a given origin. The methodology consists of two stages (Figure~\ref{fig:methodology}): i) performance estimation, which collects forecasting errors across origins in the source domain; and ii) metalearning, which models error percentiles as a function of structural features extracted from recent lags. By targeting a scale-invariant statistic (percentile rank) and using domain-agnostic inputs (time series features), the resulting rejection mechanism is designed to transfer effectively across domains and methods.

\subsection{Rolling Origin Performance Estimation}
\label{sec:performance}

The first stage of the methodology involves collecting performance estimates for metalearning. We accomplish this by conducting cross-validation over the source domain using a rolling origin evaluation procedure~\cite{hewamalage2023forecast} (also referred to as time series cross-validation~\cite{hyndman2018forecasting}). For each forecast origin $t$ (Section~\ref{sec:forecast}), the forecasting model $f_\theta$ is trained on all observations up to $t$ and is used to predict the next $h$ steps. The horizon error is computed as:
\begin{equation}
e_{i,t} = \text{loss}\bigl(\hat{Y}^{i}_{t+1:t+h},\, Y^{i}_{t+1:t+h}\bigr),
\end{equation}
where $f_\theta(X^{i}_t) = \hat{Y}^{i}_{t+1:t+h}$ is the forecast and $Y^{i}_{t+1:t+h}$ is the ground truth. Repeating this over all available series yields the set of performance estimates $\mathcal{E} = \{ e_{i,t} : i = 1,\ldots,n,\; t \in T_i \}$, where $T_i$ denotes the valid origins for series $i$.

\subsection{Metamodel Training}

\subsubsection{Meta-Targets.} 
Raw performance estimates depend on problem complexity and data scale, making them unsuitable as regression targets across heterogeneous series. Instead, we transform each error into an empirical percentile rank computed within its own series:
\begin{equation}
u_{i,t} =
\frac{\#\{j \in T_i : j \neq t,\ e_{i,j} < e_{i,t}\}}{|T_i|+1}.
\end{equation}
where $T_i$ is the set of valid origins for series $i$. This per-series normalization maps errors to a bounded target in $(0,1)$, yielding a scale-invariant score that enables the metamodel to learn transferable rejection criteria across domains.

\subsubsection{Meta-Predictors.}
The metamodel input is derived from the lags $X^{i}_t \in \mathbb{R}^p$ preceding each forecast origin. Each lag window is transformed into a feature vector $m_{i,t} = g(X^{i}_t)$, where $g(\cdot)$ is a feature extractor that summarizes structural characteristics such as trend, seasonality, and complexity~\cite{barandas2020tsfel,cerqueira2024vest}. The full set of meta-features is $\mathcal{M} = \{ m_{i,t} : i = 1,\ldots,n,\; t \in T_i \}$.

The metamodel thus learns a mapping from structural features $m_{i,t}$ to error percentiles $u_{i,t}$. Because both inputs and targets are domain-agnostic (features describe local patterns rather than absolute values, and percentiles are normalized within each series), the learned rejection function transfers across datasets without explicitly requiring retraining.

\subsection{Inference}

At inference, the metamodel scores each forecast origin before the forecast is issued. Given the lag window $X^{i}_t$, the meta-features $m_{i,t} = g(X^{i}_t)$ are computed and passed to the metamodel, which outputs a predicted error percentile $\hat{u}_{i,t}$. Because this score depends only on the input lags, not on the forecast itself, the rejection decision can be made before invoking the forecasting model, enabling pre-forecast screening, in contrast to other approaches.

The rejection rule is defined by a threshold $q \in (0,1)$ corresponding to the target coverage level: origins with $\hat{u}_{i,t} \geq q$ are rejected, and the remaining forecasts are issued. At coverage $q$, a fraction of approximately $q$ forecasts are accepted, with the rejected corresponding to those predicted to incur the largest errors.

The metamodel can be applied in two modes: (i) \emph{zero-shot}, where the source-trained metamodel is applied directly to a new domain without modification, and (ii) \emph{domain-adapted}, where the metamodel is fine-tuned on a small subset of labeled data from the target domain before deployment. The zero-shot setting tests whether the scale-invariant design transfers out of the box, and the adapted setting tests whether light supervision improves rejection quality when past target data is available.

\section{Experimental Setup}

The experiments address three research questions:

\begin{itemize}

    \item \textbf{RQ1:} Can the proposed metamodel accurately rank forecast origins by expected error, and does rejecting those origins predicted to incur large errors improve average forecasting accuracy?
    
    \item \textbf{RQ2:} Does the proposed approach outperform uncertainty-based and residual-based rejection baselines in both ranking quality and downstream forecasting accuracy?
    
    \item \textbf{RQ3:} Does the scale-invariant design enable effective transfer across domains, and how does optional domain adaptation affect performance?
    
\end{itemize}

\subsection{Datasets and Evaluation Protocol}
\label{sec:datasets}


Experiments use the widely employed M3~\cite{makridakis2000m3}, M1~\cite{makridakis1982accuracy}, and Tourism~\cite{athanasopoulos2011tourism} univariate time series collections, restricted to monthly (M) and quarterly (Q) frequencies (Table~\ref{tab:dataset_summary}). Each experiment pairs a source dataset for in-domain evaluation with a target dataset for transfer evaluation: M3$\rightarrow$M1 and M1$\rightarrow$Tourism. The forecast horizon $h$ is set to 6 for monthly data and 4 for quarterly data, with a forecast input context size of $p = 2\times h$.



\begin{table}[b]
\centering
\caption{Summary statistics of the time series datasets used in the experiments.}
\label{tab:dataset_summary}
\small
\setlength{\tabcolsep}{4pt}
\begin{tabular}{lrrr}
\toprule
Dataset & Series & Avg. len. & Windows\\
\midrule
M3 Monthly        & 1428 & 117 & 26126 \\
M3 Quarterly      & 756  & 48  & 8380  \\
\midrule
Tourism Monthly   & 366  & 298 & 17750 \\
Tourism Quarterly & 427  & 99  & 9973  \\
\midrule
M1 Monthly        & 617  & 72  & 6698  \\
M1 Quarterly      & 203  & 40  & 1850  \\
\bottomrule
\end{tabular}
\end{table}

Following the rolling origin procedure described in Section~\ref{sec:performance}, each series yields multiple forecast origins. To ensure the forecasting model has sufficient training history at each origin, only the second half of each series is used for evaluation; the first half serves exclusively as warm-up data for model fitting.

\begin{figure}[t]
    \centering
    \includegraphics[width=\linewidth]{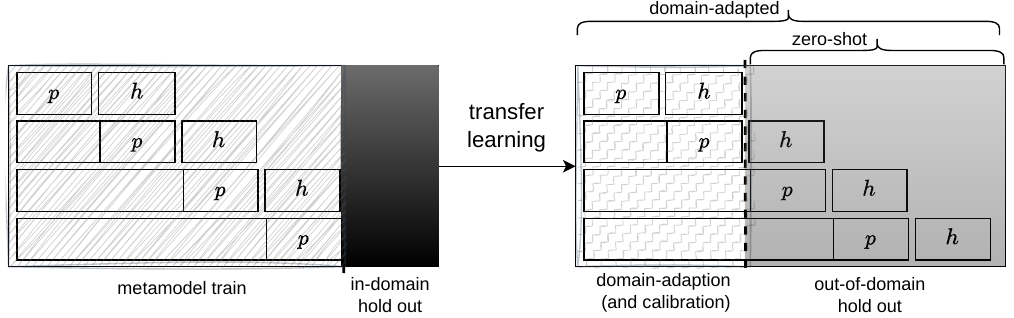}
    \caption{Evaluation protocol. \textbf{Left (source domain):} the metamodel is trained on early origins (hatched) and evaluated on the final held-out horizon (solid). \textbf{Right (target domain):} zero-shot evaluation uses only the out-of-domain holdout (solid); domain-adapted evaluation additionally fine-tunes on early target origins (hatched), which may include a calibration subset for threshold selection.}
    \label{fig:experiments}
\end{figure}

Figure~\ref{fig:experiments} illustrates the evaluation protocol. In the source domain (left), the in-domain holdout consists of the last $h$ observations of each series, corresponding to the final forecast horizon (solid region). All prior kept origins form the meta-training set (hatched region). In the target domain (right), the latest origins are held out for assessing transferability (solid region). In the zero-shot setting, the source-trained metamodel is applied directly to this holdout. In the domain-adapted setting, earlier target origins (hatched region) are used to fine-tune (i.e., domain-adapt) the metamodel. This adaptation set can be further split to reserve a calibration subset for setting the rejection threshold $q$.


\subsection{Models}
\label{sec:models}

We use two deep learning architectures for the forecasting models: \texttt{KAN}~\cite{liu2025kan}, based on Kolmogorov-Arnold Networks with learnable edge activation functions, and \texttt{NHITS}~\cite{challu2023nhits}, which employs multi-rate signal processing and hierarchical interpolation for long-horizon forecasting. Both are implemented via \texttt{NeuralForecast}\footnote{\href{https://nixtlaverse.nixtla.io/neuralforecast/models.html}{https://nixtlaverse.nixtla.io/neuralforecast/models.html}} with automatic hyperparameter tuning on 10 random search trials. Errors $e_{i,t}$ are computed using sMAPE, following its use in forecasting competitions~\cite{makridakis2000m3,makridakis2018m4}, although the proposed approach is agnostic to the metric choice.

The metamodel is a \texttt{CatBoostRegressor}~\cite{prokhorenkova2018catboost} that predicts empirical error percentiles $u_{i,t}$ from time series features $m_{i,t}$. Features are extracted using \texttt{TSFEL}~\cite{barandas2020tsfel}, comprising statistical, temporal, and spectral descriptors of the lag window preceding each forecast origin.\footnote{\href{https://tsfel.readthedocs.io/en/latest/descriptions/feature_list.html}{https://tsfel.readthedocs.io/en/latest/descriptions/feature\_list.html}} Missing feature values are imputed with the median. Hyperparameters (tree depth, learning rate, $L_2$ regularization, bagging temperature, border count, boosting iterations) are selected via random search over 30 configurations, using grouped cross-validation to keep all origins from the same series within the same fold and prevent leakage.

\subsection{Baselines}
\label{sec:baselines}

We compare the metamodel against the following baselines:

\begin{itemize}
    \item \textbf{PI-width}: Inspired by conformal prediction~\cite{shafer2008tutorial}, this baseline uses prediction interval width as the rejection score. Residuals from recent origins are normalized by the median absolute deviation (MAD), and the 90\% quantile defines a per-step interval. The mean interval width across the forecasting horizon serves as the rejection score.

    \item \textbf{Residual-scale}: Measures the variability of the recent loss by computing the MAD of sMAPE values from the most recent forecast origins. Higher variability indicates less stable predictions.

    \item \textbf{Residual-variance}: Pools pointwise residuals from recent origins to estimate variance, then converts this to a 90\% Student-$t$ interval width. Larger widths indicate greater uncertainty.

    \item \textbf{Random}: Rejects windows uniformly at random, matching the rejection count of other methods at each coverage level. Serves as a lower bound.

    \item \textbf{Oracle}: Ranks windows by their realized sMAPE and rejects the highest-error windows first. Serves as an upper bound, representing the best achievable performance when true errors are known.
\end{itemize}

\section{Results}

We evaluate the metamodel at two levels: (i) meta-level, assessing whether predicted risk scores rank forecast origins by actual error, and (ii) base-level, measuring downstream forecasting accuracy after rejection.

\subsection{Meta-level Evaluation}

Table~\ref{tab:metamodel-ranking-both} compares the predicted large error scores against the realized errors across source (S), zero-shot transfer (Z), and domain-adapted transfer (A) settings. Metrics include Spearman's $\rho$ (rank correlation between $\hat{u}_{i,t}$ and $u_{i,t}$), AUCO, and ErrDrop (Section~\ref{sec:reject_option}). Metamodel performance is consistent across both forecasters, indicating method-agnostic behavior.

\subsubsection{In-domain Settings ($S$).} Spearman correlation ($\rho$) is consistently high (i.e., between $0.71$ and $0.90$), indicating that the metamodel correctly assigns higher scores to windows that actually lead to the largest errors. This is further supported by the small AUCO values the metamodel attains, which are closest to the oracle across all settings. The ErrDrop metric also shows that selective rejection can substantially reduce the average error on accepted forecasts at the extremes, as the metamodel is always either the best or the second-best.

\subsubsection{Transfer Settings ($Z/A$).} 
Zero-shot transfer degrades metamodel performance relative to source (S), as expected, due to domain shift, with lower $\rho$, larger AUCO, and generally smaller ErrDrop, though still outperforming most baselines. Domain adaptation using 30\% of target origins substantially recovers performance: $\rho$ and AUCO improve in all cases, and ErrDrop improves in all but one (M1-M$\rightarrow$T-M). Bold values indicate the better transfer setting.

The baselines only outperform the metamodel in isolated cases. \texttt{Residual scale} is occasionally stronger in zero-shot transfer for $\rho$ and AUCO, but these gains are small and disappear after target adaptation. \texttt{PI width} and \texttt{Error Variance} win on ErrDrop only in a few source-domain cases, indicating that they sometimes identify the most extreme errors, but do not provide reliable transfer rankings. A dash indicates that the statistic is undefined because the corresponding score is constant or fully tied over the evaluated windows.

\begin{table}[!ht]
\centering
\scriptsize
\setlength{\tabcolsep}{1pt}
\renewcommand{\arraystretch}{0.90}
\caption{Evaluation of the metamodel alongside \texttt{NHITS} and \texttt{KAN} forecasting models. Each cell reports values for source holdout (S), zero-shot transfer (Z), and domain adaptation (A). The best method in each setting is bolded, and the second best is underlined. Spearman ($\rho$) measures rank association, AUCO measures area under calibration-oracle error curve, and ErrDrop is the keep-all to most-selective risk ratio. A dash indicates that the corresponding metric is constant or fully tied over the evaluated windows.}
\label{tab:metamodel-ranking-both}
\resizebox{.95\textwidth}{!}{%
\begin{tabular}{@{}llll|c|c|c@{}}
\toprule
Source & Target & Models & Method & $\rho\uparrow$ (S; Z/A) & AUCO$\downarrow$ (S; Z/A) & ErrDrop$\uparrow$ (S; Z/A) \\
\midrule
M3-M & M1-M & KAN & \textbf{Ours} & \textbf{0.894};\; \textbf{0.628}/\textbf{0.820} & \textbf{0.007};\; \textbf{0.043}/\textbf{0.013} & \textbf{19.32};\; \textbf{10.06}/\textbf{14.23} \\
 &  &  & PI width & 0.687;\; 0.158/0.158 & 0.023;\; 0.124/0.124 & 11.50;\; 0.67/0.67 \\
 &  &  & Res. scale & 0.804;\; \underline{0.536}/\underline{0.536} & \underline{0.012};\; \underline{0.055}/\underline{0.055} & 12.38;\; \underline{1.71}/\underline{1.71} \\
 &  &  & Err. var. & \underline{0.818};\; 0.152/0.152 & 0.012;\; 0.122/0.122 & \underline{15.97};\; 0.76/0.76 \\
\cmidrule(lr){3-7}
 &  & NHITS & \textbf{Ours} & \textbf{0.899};\; \textbf{0.571}/\textbf{0.812} & \textbf{0.006};\; \textbf{0.045}/\textbf{0.013} & \textbf{28.58};\; \textbf{19.18}/\textbf{15.39} \\
 &  &  & PI width & 0.671;\; 0.110/0.110 & 0.024;\; 0.140/0.140 & 10.78;\; 0.45/0.45 \\
 &  &  & Res. scale & 0.801;\; \underline{0.536}/\underline{0.536} & 0.012;\; \underline{0.047}/\underline{0.047} & 15.92;\; \underline{1.27}/\underline{1.27} \\
 &  &  & Err. var. & \underline{0.815};\; 0.105/0.105 & \underline{0.012};\; 0.134/0.134 & \underline{19.17};\; 1.14/1.14 \\
\midrule
M3-Q & M1-Q & KAN & \textbf{Ours} & \textbf{0.750};\; \underline{0.529}/\textbf{0.755} & \textbf{0.008};\; 0.059/\textbf{0.013} & \underline{4.74};\; \textbf{9.38}/\textbf{11.77} \\
 &  &  & PI width & 0.556;\; 0.220/0.220 & 0.016;\; 0.093/0.093 & \textbf{6.24};\; 0.44/0.44 \\
 &  &  & Res. scale & 0.466;\; \textbf{0.535}/\underline{0.535} & 0.022;\; \textbf{0.056}/\underline{0.056} & 1.66;\; 0.78/0.78 \\
 &  &  & Err. var. & \underline{0.663};\; 0.311/0.311 & \underline{0.012};\; \underline{0.058}/0.058 & 4.23;\; \underline{7.73}/\underline{7.73} \\
\cmidrule(lr){3-7}
 &  & NHITS & \textbf{Ours} & \textbf{0.747};\; \underline{0.515}/\textbf{0.734} & \textbf{0.007};\; \textbf{0.037}/\textbf{0.013} & \textbf{10.54};\; \textbf{11.66}/\textbf{15.55} \\
 &  &  & PI width & 0.612;\; 0.165/0.165 & 0.013;\; 0.086/0.086 & \underline{7.75};\; 0.50/0.50 \\
 &  &  & Res. scale & 0.520;\; \textbf{0.537}/\underline{0.537} & 0.018;\; \underline{0.047}/\underline{0.047} & 1.87;\; 0.84/0.84 \\
 &  &  & Err. var. & \underline{0.695};\; 0.199/0.199 & \underline{0.010};\; 0.068/0.068 & 5.90;\; \underline{7.38}/\underline{7.38} \\
\midrule
M1-M & T-M & KAN & \textbf{Ours} & \textbf{0.752};\; \underline{0.494}/\textbf{0.717} & \textbf{0.016};\; \underline{0.058}/\textbf{0.035} & \textbf{4.99};\; \textbf{2.58}/\underline{2.44} \\
 &  &  & PI width & 0.213;\; -0.272/-0.272 & 0.106;\; 0.180/0.180 & 0.27;\; 0.50/0.50 \\
 &  &  & Res. scale & \underline{0.485};\; \textbf{0.519}/\underline{0.519} & \underline{0.036};\; \textbf{0.053}/\underline{0.053} & \underline{2.74};\; \underline{2.44}/\textbf{2.44} \\
 &  &  & Err. var. & 0.210;\; -0.316/-0.316 & 0.092;\; 0.186/0.186 & 1.21;\; 0.47/0.47 \\
\cmidrule(lr){3-7}
 &  & NHITS & \textbf{Ours} & \textbf{0.766};\; \textbf{0.559}/\textbf{0.741} & \textbf{0.014};\; \textbf{0.054}/\textbf{0.033} & \textbf{11.01};\; \textbf{2.41}/\textbf{2.91} \\
 &  &  & PI width & 0.209;\; -0.251/-0.251 & 0.087;\; 0.188/0.188 & 0.35;\; 0.52/0.52 \\
 &  &  & Res. scale & \underline{0.482};\; \underline{0.473}/\underline{0.473} & \underline{0.038};\; \underline{0.067}/\underline{0.067} & \underline{1.92};\; \underline{1.58}/\underline{1.58} \\
 &  &  & Err. var. & 0.204;\; -0.288/-0.288 & 0.082;\; 0.195/0.195 & 0.62;\; 0.47/0.47 \\
\midrule
M1-Q & T-Q & KAN & \textbf{Ours} & \textbf{0.710};\; \textbf{0.544}/\textbf{0.764} & \textbf{0.015};\; \textbf{0.053}/\textbf{0.030} & \underline{2.88};\; \textbf{2.92}/\textbf{3.28} \\
 &  &  & PI width & 0.247;\; -0.016/-0.016 & 0.059;\; 0.125/0.125 & 2.21;\; 1.33/1.33 \\
 &  &  & Res. scale & --;\; \underline{0.255}/\underline{0.255} & \underline{0.049};\; \underline{0.099}/\underline{0.099} & 1.07;\; 1.64/1.64 \\
 &  &  & Err. var. & \underline{0.257};\; -0.014/-0.014 & 0.062;\; 0.123/0.123 & \textbf{3.08};\; \underline{1.80}/\underline{1.80} \\
\cmidrule(lr){3-7}
 &  & NHITS & \textbf{Ours} & \textbf{0.733};\; \textbf{0.525}/\textbf{0.751} & \textbf{0.013};\; \textbf{0.060}/\textbf{0.035} & \textbf{10.85};\; \textbf{3.05}/\textbf{2.76} \\
 &  &  & PI width & 0.247;\; 0.004/0.004 & 0.058;\; 0.136/0.136 & 2.14;\; 1.28/1.28 \\
 &  &  & Res. scale & --;\; \underline{0.287}/\underline{0.287} & \underline{0.050};\; \underline{0.096}/\underline{0.096} & 0.98;\; 1.48/1.48 \\
 &  &  & Err. var. & \underline{0.269};\; 0.008/0.008 & 0.058;\; 0.135/0.135 & \underline{2.81};\; \underline{1.58}/\underline{1.58} \\
\bottomrule
\end{tabular}
}
\end{table}

\begin{figure}[h]
    \centering
    \includegraphics[width=1\linewidth]{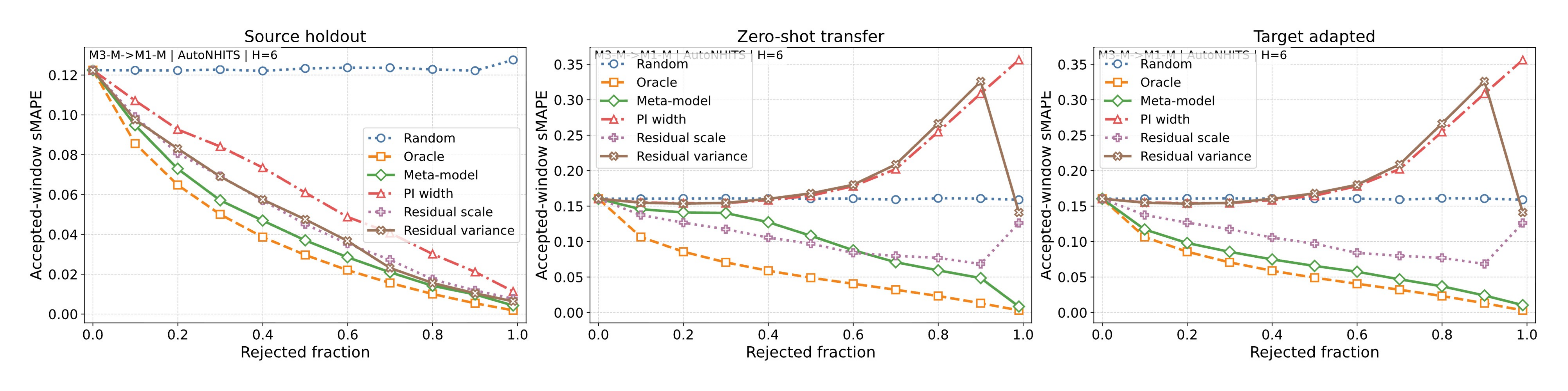}
    \caption{Risk-Coverage plots for the \texttt{NHITS} forecaster, using M3 Monthly as the source domain and M1 Monthly as the target domain. The metamodel (green diamond) is closest to the oracle in the source domain across most coverage levels, and it consistently becomes the closest after domain adaptation in the target domain.}
    \label{fig:risk-coverage}
\end{figure}

Figure~\ref{fig:risk-coverage} shows one example using M3 Monthly as the source and M1 Monthly as the target in risk-coverage curves. In the source domain (left), the metamodel (green diamond) is consistently closest to the oracle (orange square) across rejection levels, indicating that it is generally the best at predicting high-error forecasts. The other baselines, even though they leverage the actual residuals to compute risk scores, only become competitive at higher levels. As expected, the random baseline remains relatively unaltered because it leverages no information about which forecasts to reject. In the transfer domain plots (center and right), \texttt{PI width} and \texttt{Residual-variance} severely underperform, being worse than the random baseline around the $0.5$ level. This indicates that past error variance becomes progressively decoupled from transfer learning forecasting risk as the rejection fraction increases. Only \texttt{Residual-scale} approximates the metamodel, though it becomes unreliable at the largest rejections.

However, applying domain adaptation (right) immediately approximates the metamodel to the oracle upper bound, demonstrating that it is crucial for robustness when domain shift is present.

\subsection{Base-level Evaluation}

\begin{table}[h]
\centering
\scriptsize
\setlength{\tabcolsep}{1.2pt}
\caption{Base-level results on the transfer target datasets across $q$ rejection fractions. Keep all is target-domain sMAPE without rejection. Baseline lists the abstention methods used. Each $q$ reports sMAPE on kept forecasts by each method. Avg. gap reports the average distance to the oracle across $q$. Lower values and smaller gaps are better.}
\label{tab:online-reject-option-gap-to-oracle}
\begin{tabular}{@{}llrlcccccc@{}}
\toprule
Target & Models & Keep all$\downarrow$ & Baseline & $q=0.05$ & $q=0.10$ & $q=0.20$ & $q=0.30$ & $q=0.40$ & Avg. gap$\downarrow$ \\
\midrule
T-M & KAN & 0.288 & \textbf{Ours} & \underline{0.278} & \textbf{0.267} & \textbf{0.249} & \textbf{0.233} & \textbf{0.215} & \textbf{0.033} \\
 &  &  & PI width & 0.287 & 0.291 & 0.304 & 0.313 & 0.325 & 0.089 \\
 &  &  & Res. scale & \textbf{0.278} & \underline{0.268} & \underline{0.255} & \underline{0.241} & \underline{0.228} & \underline{0.039} \\
 &  &  & Err. var. & 0.290 & 0.295 & 0.309 & 0.316 & 0.329 & 0.093 \\
\cmidrule(lr){4-10}
 &  &  & Oracle & 0.259 & 0.241 & 0.214 & 0.191 & 0.172 & 0.000 \\
\cmidrule(lr){2-10}
 & NHITS & 0.307 & \textbf{Ours} & \textbf{0.295} & \textbf{0.284} & \textbf{0.264} & \textbf{0.244} & \textbf{0.226} & \textbf{0.034} \\
 &  &  & PI width & 0.304 & 0.310 & 0.322 & 0.330 & 0.343 & 0.094 \\
 &  &  & Res. scale & \underline{0.298} & \underline{0.290} & \underline{0.277} & \underline{0.262} & \underline{0.249} & \underline{0.047} \\
 &  &  & Err. var. & 0.307 & 0.313 & 0.326 & 0.335 & 0.345 & 0.097 \\
\cmidrule(lr){4-10}
 &  &  & Oracle & 0.275 & 0.256 & 0.226 & 0.203 & 0.182 & 0.000 \\
\midrule
T-Q & KAN & 0.272 & \textbf{Ours} & \textbf{0.239} & \textbf{0.229} & \textbf{0.210} & \textbf{0.193} & \textbf{0.180} & \textbf{0.021} \\
 &  &  & PI width & 0.260 & 0.262 & 0.261 & 0.260 & 0.259 & 0.071 \\
 &  &  & Res. scale & 0.265 & \underline{0.258} & \underline{0.247} & \underline{0.241} & \underline{0.236} & \underline{0.060} \\
 &  &  & Err. var. & \underline{0.259} & 0.260 & 0.260 & 0.259 & 0.256 & 0.069 \\
\cmidrule(lr){4-10}
 &  &  & Oracle & 0.235 & 0.217 & 0.188 & 0.164 & 0.144 & 0.000 \\
\cmidrule(lr){2-10}
 & NHITS & 0.304 & \textbf{Ours} & \textbf{0.278} & \textbf{0.266} & \textbf{0.245} & \textbf{0.229} & \textbf{0.211} & \textbf{0.025} \\
 &  &  & PI width & 0.295 & 0.295 & 0.295 & 0.293 & 0.293 & 0.073 \\
 &  &  & Res. scale & \underline{0.291} & \underline{0.281} & \underline{0.267} & \underline{0.259} & \underline{0.253} & \underline{0.050} \\
 &  &  & Err. var. & 0.292 & 0.294 & 0.294 & 0.293 & 0.290 & 0.072 \\
\cmidrule(lr){4-10}
 &  &  & Oracle & 0.268 & 0.250 & 0.220 & 0.195 & 0.171 & 0.000 \\
\midrule
M1-Q & KAN & 0.143 & \textbf{Ours} & \textbf{0.124} & \underline{0.115} & \textbf{0.061} & \textbf{0.050} & \textbf{0.042} & \textbf{0.019} \\
 &  &  & PI width & 0.139 & 0.134 & 0.133 & 0.131 & 0.131 & 0.074 \\
 &  &  & Res. scale & \underline{0.125} & \textbf{0.110} & \underline{0.096} & \underline{0.091} & \underline{0.084} & \underline{0.042} \\
 &  &  & Err. var. & 0.139 & 0.138 & 0.140 & 0.144 & 0.139 & 0.081 \\
\cmidrule(lr){4-10}
 &  &  & Oracle & 0.095 & 0.074 & 0.054 & 0.041 & 0.033 & 0.000 \\
\cmidrule(lr){2-10}
 & NHITS & 0.120 & \textbf{Ours} & \underline{0.103} & \underline{0.098} & \textbf{0.060} & \textbf{0.048} & \textbf{0.040} & \textbf{0.016} \\
 &  &  & PI width & 0.113 & 0.109 & 0.107 & 0.105 & 0.103 & 0.053 \\
 &  &  & Res. scale & \textbf{0.098} & \textbf{0.087} & \underline{0.069} & \underline{0.067} & \underline{0.067} & \underline{0.023} \\
 &  &  & Err. var. & 0.113 & 0.112 & 0.109 & 0.112 & 0.110 & 0.057 \\
\cmidrule(lr){4-10}
 &  &  & Oracle & 0.084 & 0.067 & 0.050 & 0.039 & 0.031 & 0.000 \\
\bottomrule
\end{tabular}
\end{table}

We simulate sequential deployment in the target domain to assess the impact of the metamodel at the base level. The metamodel is domain-adapted using the earlier target windows, and an abstention threshold is computed for several fixed rejection fractions $q$ on the calibration windows. It is then applied to the held-out windows to assess whether the metamodel effectively reduces accepted-forecast error as the thresholds increase (i.e., risk, as defined in Section~\ref{sec:reject_option}). 

Table~\ref{tab:online-reject-option-gap-to-oracle} showcases results for both forecasters on all datasets, when no forecast is rejected, and across increasing rejection thresholds $q$, for all methods. We also show the average distance to the oracle upper bound at the end, representing the error if we actually rejected the top-$q$ worst windows at each reported level. Once again, the similar improvements attained across both forecasters indicate that the metamodel is model-agnostic.

We consistently improve over the \texttt{Keep all} baseline at each level, and improve over the other baselines in most (in bold). This is also evidenced by the average gap to the oracle across $q$, at which our method consistently achieves the lowest. Nonetheless, when our method is not the best, usually at lower $q$ values, it is the second best, denoted by an underline. Moreover, a series-level bootstrap confirmed that the metamodel is significantly closer to the oracle than all baselines in all settings, except \texttt{Residual-scale} for M1-Q/NHITS ($p=0.191$).

\begin{figure}[t]
    \centering
    \includegraphics[width=1\linewidth]{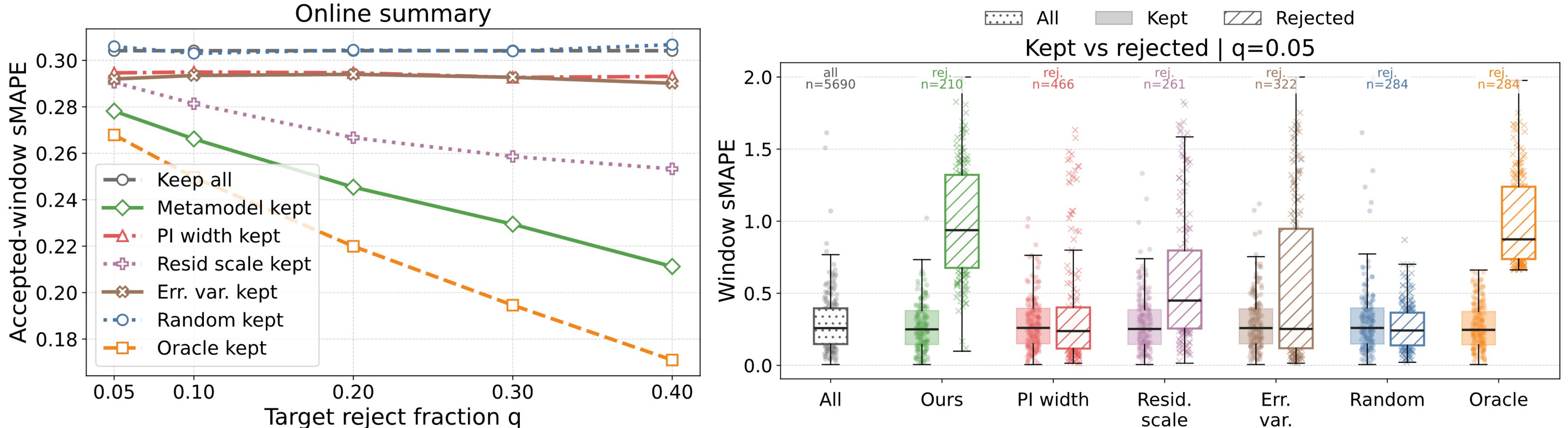}
    \caption{Impact of the selective forecasting mechanism at base level across $q$ thresholds, from M1-T to T-Q. On the left, even at higher $q$ (more rejections), the metamodel remains close to the oracle. The right plot shows the distributions of all errors across windows, along with those kept and rejected at $q$ = 0.05 for each method.}
    \label{fig:online}
\end{figure}

Figure~\ref{fig:online} showcases whether abstaining from forecasts rejected by the metamodel improves the average error on retained forecasts. The left plot reports the mean sMAPE across accepted forecasts for several rejection levels $q$ and all methods. The no-abstention baseline (keep all) remains essentially unchanged, as does random rejection. Although the \texttt{PI width} and \texttt{Error variance} baselines slightly improve over random and no abstention, both remain constant across $q$, showcasing their limitations on attainable improvement. Both the \texttt{Residual scale} baseline and the metamodel rejection curves decrease as $q$ increases, with the latter decreasing more prominently. The persistent gap between the metamodel and the oracle, which is nearly half of the \texttt{Residual scale} gap, indicates that, even across rejection thresholds, our method remains the most reliable. 

The right plot compares the error distributions for all, the retained, and the rejected forecasts by each method at the strictest $q$ value ($0.05$). It is clear that forecasts rejected by the metamodel, \texttt{Residual scale}, and \texttt{Error variance} exhibit larger errors than those retained, indicating that they concentrate large errors in the rejected subset. Our method (green) is particularly effective, as its distribution of rejected forecasts is closest to the oracle distribution (orange). 

\section{Discussion, Limitations, and Conclusion}

We present a selective forecasting framework that predicts empirical error percentiles from structural features of recent observations, enabling abstention before the forecast is issued. The scale-invariant targets and domain-agnostic features allow the mechanism to transfer across heterogeneous domains.

\paragraph{Research Questions.}

Results show that the metamodel can identify instances associated with large forecasting errors by predicting empirical error percentiles, and that using these to selectively forecast improves downstream performance on the retained windows (\textbf{RQ1}). This indicates that the extracted meta-features contain useful information about forecasting difficulty, even before the forecast is produced. The fact that the metamodel learns patterns from several time series via cross-learning may explain its ability to capture general conditions for forecastability across heterogeneous windows.

The proposed method is competitive with uncertainty-based and residual-based post-hoc baselines for identifying windows associated with large forecasting errors, both in-domain and in transfer learning settings (\textbf{RQ2}). Transfer learning results also show that the metamodel is robust across most settings, particularly when domain adaptation is applied. This indicates that forecastability information can generalize across datasets, although distribution shift can reduce metamodel performance when source meta-features are not representative of the target domain. Unlike methods that employ prediction intervals or residual behavior, which are only available after the forecast is produced and may require direct knowledge of recent errors, the proposed method operates \textit{ex ante}. It only uses descriptors extracted from lags, making it suitable as a pre-forecast triage mechanism.

Results further show that abstaining from forecasts rejected by the metamodel improves average predictive performance on the retained forecasts, producing competitive selective-risk trade-offs relative to uncertainty-based and residual-based post-hoc baselines in transfer learning settings (\textbf{RQ3}). This suggests that the rejected windows correspond to conditions under which the forecaster is more likely to incur large errors, while the retained forecasts are more closely aligned with previously observed predictable conditions.

\paragraph{Limitations.}

The method relies heavily on the representativeness of the meta-training data, as domains with shorter series can make the rejection mechanism less reliable. Moreover, the method estimates relative forecasting risk rather than calibrated probabilistic guarantees, and thus does not provide formal quantifications of uncertainty. Moreover, given the novelty of the approach, all comparative baselines are based on recently incurred residuals, which makes them harder to beat but may not be representative of other, narrower approaches in the literature (e.g., those applicable only to time series foundation models). 

Results suggest that forecastability can be treated as a structural property of recent observations, enabling risk-aware decisions before the forecast horizon is realized. Future directions include integrating the rejection mechanism with conformal prediction to achieve calibrated coverage guarantees, extending to multivariate and irregular time series, and exploring joint training of the forecaster and the rejection model.

\begin{credits}
\subsubsection{\ackname} This work was partially funded by projects AISym4Med (101095387) supported by Horizon Europe Cluster 1: Health, ConnectedHealth (n.º 46858), supported by Competitiveness and Internationalisation Operational Programme (POCI) and Lisbon Regional Operational Programme (LISBOA 2020), under the PORTUGAL 2020 Partnership Agreement, through the European Regional Development Fund (ERDF) and Agenda “Center for Responsible AI”, nr. C645008882-00000055, investment project nr. 62, financed by the Recovery and Resilience Plan (PRR) and by European Union -  NextGeneration EU, and also by FCT plurianual funding for 2020-2023 of LIACC (UIDB/00027/2020 UIDP/00027/2020);
\end{credits}

\bibliographystyle{splncs04}
\bibliography{mybibliography}

\end{document}